\documentclass{article} 
\usepackage{iclr2023_conference,times}

\usepackage{amsmath,amsfonts,bm}

\def\eqref#1{equation~\ref{#1}}

\def\1{\bm{1}}

\DeclareMathAlphabet{\mathsfit}{\encodingdefault}{\sfdefault}{m}{sl}
\SetMathAlphabet{\mathsfit}{bold}{\encodingdefault}{\sfdefault}{bx}{n}

\usepackage{hyperref}
\usepackage{url}
\usepackage{graphicx}
\usepackage{ulem}

\title{DocPrompt: Large-scale continue pretrain for zero-shot and few-shot document question answering}

\author{Sijin Wu, Dan Zhang, Teng Hu \& Shikun Feng\\
Department of Nature Language Processing\\
Baidu Inc.\\
\texttt{\{wusijin,zhangdan20,huteng,fengshikun01\}@baidu.com} \\
}

\iclrfinalcopy 
\begin{document}

\maketitle

\begin{abstract}
In this paper, we propose Docprompt for document question answering tasks with powerful zero-shot and few-shot performance. We proposed a novel weakly supervised data generation method, a novel multl-stage training method and a novel understanding model \& generation model ensemble method. We achieved state-of-the-art performance on 4 document question answering tasks. 
This method greatly improves the delivery efficiency and model performance of document question answering customer projects, reducing annotation costs and labor costs. Our demo can be found at https://huggingface.co/spaces/PaddlePaddle/ERNIE-Layout.
\end{abstract}

\section{Introduction}
\label{gen_inst}
In business scenarios across multiple industries, the proportion of document input information is gradually increasing. For example, the shipping Customs declaration and aircraft maintenance order of the shipping industry, the prospectus of the financial industry, the statements and meeting minutes of the energy and power industry, etc. Therefore, utilizing document question and answer models to efficiently, universally, and accurately extract specified information from documents has become an important means to assist users in their work and information processing.


\begin{figure}[h]
\begin{center}
\includegraphics[scale=0.6]{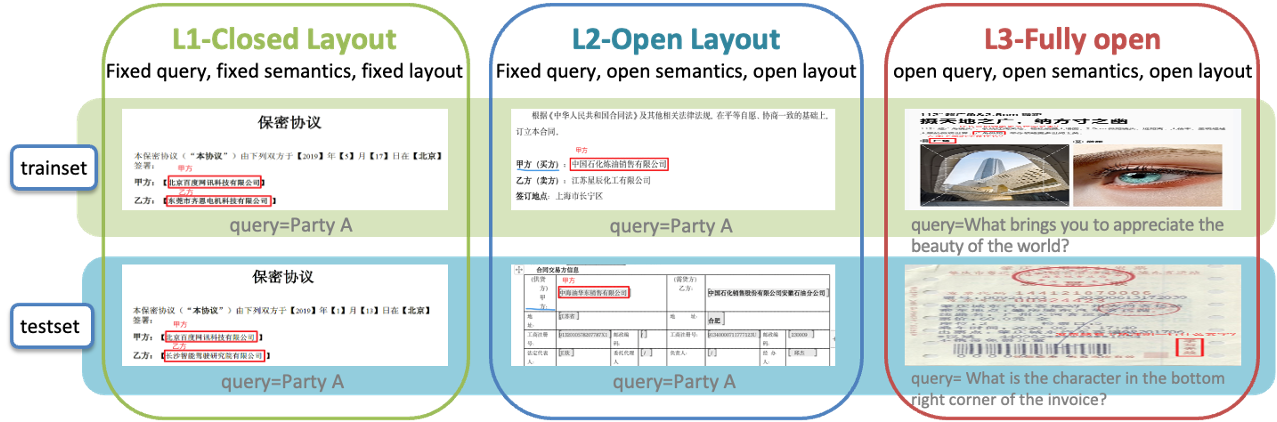}
\end{center}
\caption{Task Defination}
\end{figure}

In a real user business scenario, the manual annotation of Natural language processing tasks is very expensive because the annotator needs to understand the text content and have a certain understanding of business knowledge. Therefore, the generalization of the model and the ability to learn from zero or few samples are very important - this determines whether our model can low-cost mass deliver to multiple customers in order to achieve maximum commercial value.


We divided the document question answering task into three levels based on the required model capability level and the degree of generalization of tasks. We called it L1, L2, and L3.

In order to improve the zero-shot and few-shot capabilities of the model as well as improve the generalization ability of the model, and let the model solve L3 level document question answering tasks. We need to enable the model to learn all prior knowledge for question answering task, which including text, layout, and images. For example, "November 2022" is a date, and "Obama" is a personal name. Text of the same type in the table will be arranged in the same column, and the first row in the middle of the page is the title. The text in the columns needs to be read column by column from left to right, etc.


We proposes a large-scale continue pretrain method for document question answering, which has strong generalization and strong ability for question answering tasks in scenarios with zero or few samples. And therefore, it has the characteristic of being ready to use out of the box, greatly reducing the labor and time costs of privatization. Our demo is available at huggingface space: https://huggingface.co/spaces/PaddlePaddle/ERNIE-Layout. Feel free to have a try.


\section{Related work}
\subsection{Document Understanding pretrain Models}
LayoutLM\cite{Xu_2020}, LayoutLMv2\cite{xu2022layoutlmv2}, LayoutXLM\cite{xu2021layoutxlm} and LayoutLMv3\cite{huang2022layoutlmv3}, is a series of document understanding pretrain models. In LayoutLM, they used the layout bounding box to cut the image into pieces and then added them into the corresponding text representation. In LayoutLMv2, the authors cut the images into pieces and concatenated them at the end of the text tokens. In LayoutXLM, It has achieved the integration of multiple languages. In LayoutLMv3, they used the linear layers as the image feature extractor, which bringing better flexibility.


Ernie-layout\cite{peng2022ernielayout} is a document pretraining model proposed by Baidu and also a model used by Docprompt for hot start. It proposes innovative attention mechanisms and various pretraining tasks. ERNIE-MMLayout\cite{wang2022erniemmlayout} considers the importance of different granularity element (fragment, paragraph) relationships for document understanding. Based on ERNIE-Layout, a multi-granularity and multi-modal Transformer layer based on GNN is introduced to achieve Document Graph Aggregation representation.

LiLT \citep{wangLiLT2022} focuses on the ability of cross-lingual zero-shot transfer learning and proposes the first language-independent layout Transformer based on the decoupled text and layout flow with a bi-directional attention complementation mechanism.

MarkupLM\cite{li2022markuplm} is a pretrain model for webpage, it uesed the DOM Tree and XPath information to boosting the performance of the model.


Docprompt cannot compare with these pretrained models in terms of universality, but its finetune, zero-shot, and few-shot performance on document question answering tasks far surpass them.

\subsection{Document data generation}

\citet{kimOCRfreeDocumentUnderstanding2022} propose the SynthDoG to generate synthetic document images with the text from Wikipedia for unsupervised pre-training. They used multi-layer render method, which can better imitate the real picture.
However the generated document images are not annotated and the layout is rule-based.
Moreover, the SynthDoG renders the document images ignoring the relationship between layout and semantics.
They used these data to train a end-to-end OCR-free VDU model named Donut.

\section{Docprompt}
\subsection{Pretrain data generation}


In order to enable the model to learn more comprehensive and rich common sense knowledge and document layout knowledge, and enhance its ability related to document  question answering, we need to search for as much document question answering data as possible to continuously pretrain our model.


Due to the high annotation cost of document question answering data, we chose the remote supervision method to construct weakly supervised document question answering training data.


We use data sources such as Chinese and English Wikipedia and Baidu Baike. For each entity (such as someone's introduction) and their introduction, we use structured information (such as age: 19 years old; place of birth: Beijing) to match in the original text of the encyclopedia webpage. If the value of a structured information can be found, it is considered that a question answering task data has been successfully matched.


Through the above scheme, we have obtained a significant amount of weakly supervised text question answering task training data. Next, we need to convert it into data for document question answering tasks. 

\begin{figure}[h]
\begin{center}
\includegraphics[scale=0.9]{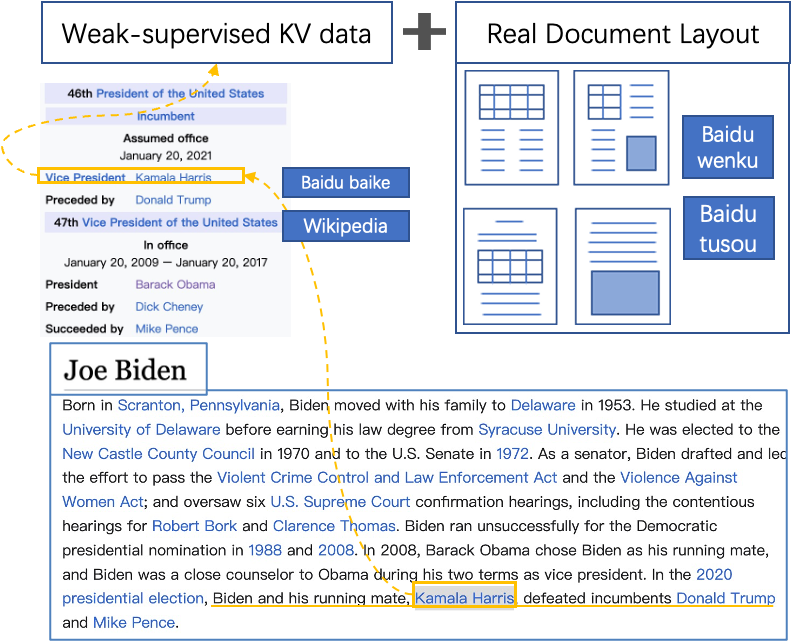}
\end{center}
\caption{Weakly supervised data construction}
\end{figure}

The model needs to learn as much layout knowledge as possible from a large number of real documents, so it is best to choose real document layouts instead of forging layout information like other works.


Therefore, we obtained a large number of unsupervised documents from Baidu Library and sent them to OCR for parsing. From this, the positional information of each word and each document element (such as tables, images, titles, etc.) in these documents was obtained.

Next, we will fill in the weakly supervised question answering data obtained in the previous step verbatim in these document layouts. Combine them into training data with weakly supervised question answering annotations and real document layout.


We will initialize a blank canvas and then render the document content onto the canvas according to the layout coordinates. This results in an image of weakly supervised data.

Through this method, we obtained pretraining data of document question answering tasks for hundreds of millions.

\subsection{Model Architecture}


Overall, we adopted Ernie-layout\cite{peng2022ernielayout} as the pretraining model and added a linear layer at the top for sequence labeling task, predicting the labels of each token.


We have added an input feature called task id to unify pure text input and document input. The layout coordinates corresponding to plain text input are all 0.

\subsubsection{Multiple-style decode}

\begin{figure}[h]
\begin{center}
\includegraphics[scale=0.7]{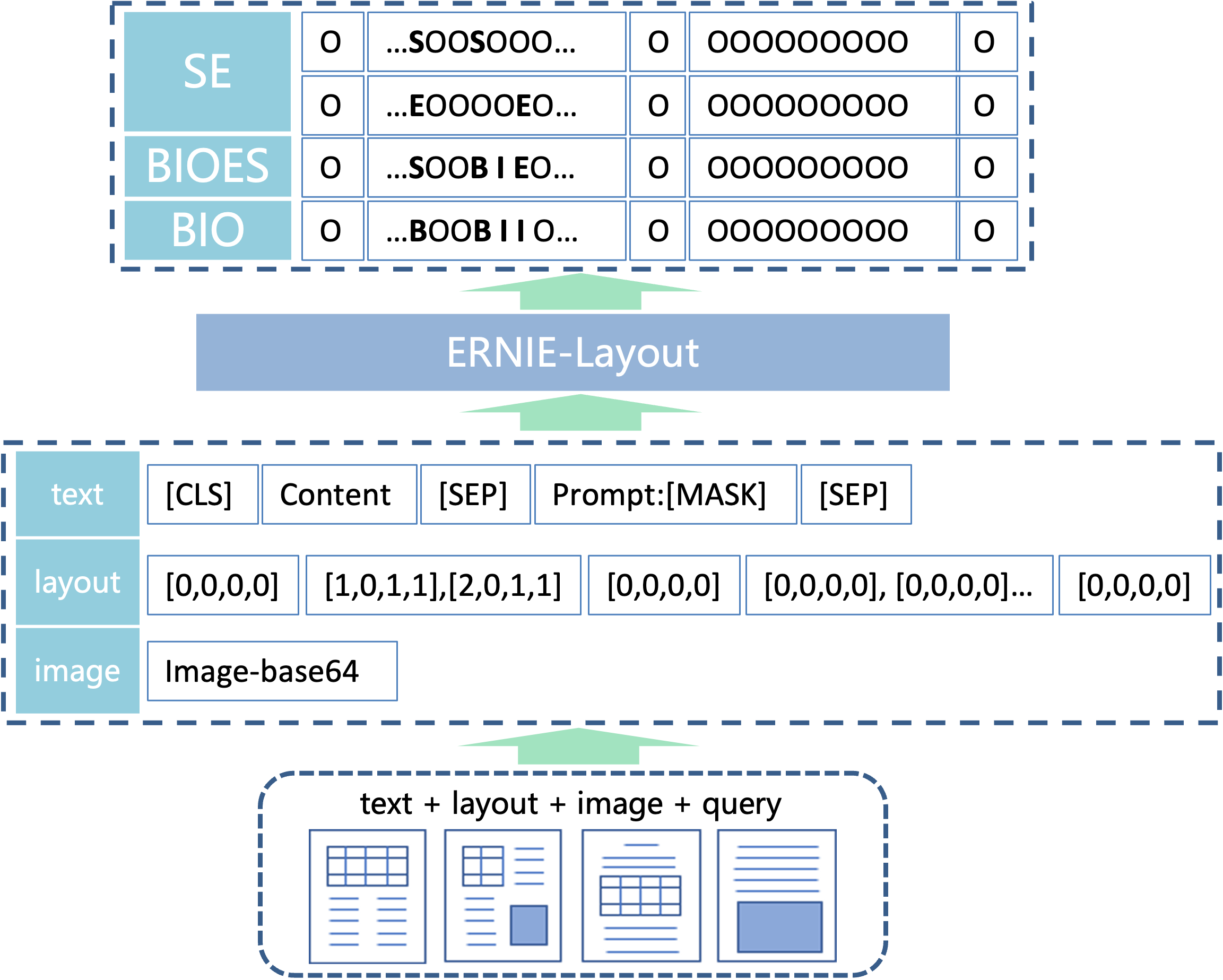}
\end{center}
\caption{Multiple-style decode}
\end{figure}


We adopt the MRC task format, concatenating document content and questions one by one, and then inputting them into the model to predict the results.


In the sequence annotation task, previous work has proposed multiple decoding methods, including BIO, BIOES, SE (start end), etc. We added multiple linear layers to predict the BIO, BIOES, and SE labels of each token, and add the 3 kinds of loss together. For the inference stage, 3 linear layers inference separately and perform Viterbi decoding separately. Finally, the predicted results of the three decoding methods were fused using voting rules.

\subsubsection{Multiple-task learning}


A new document multi-task learning scheme is adopted, which can complete most of the document tasks with a unified model and make full use of GPU resources and data resources.


\begin{figure}[h]
\begin{center}
\includegraphics[scale=0.7]{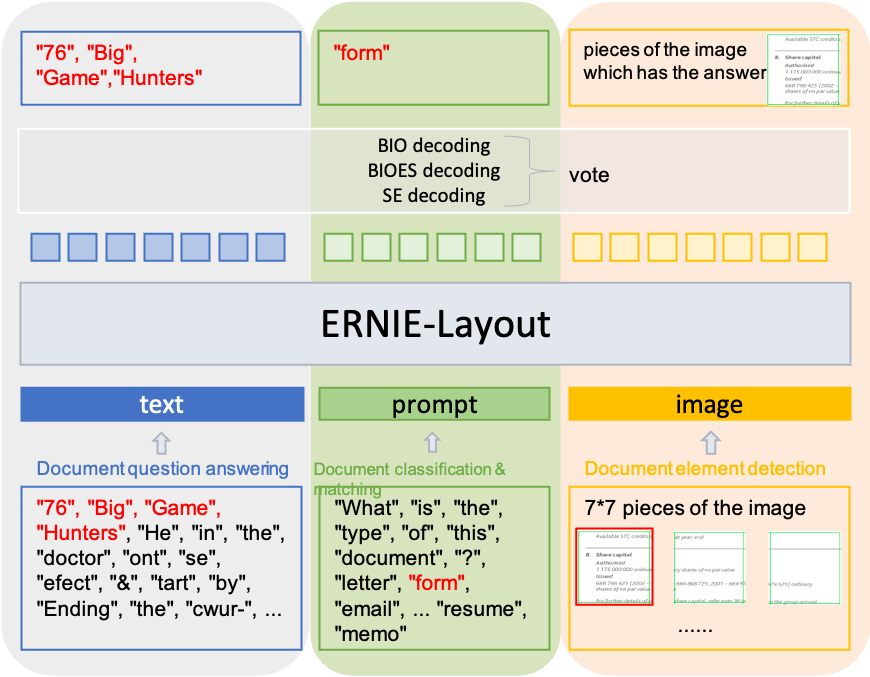}
\end{center}
\caption{Docprompt model structure and multi-task learning method}
\end{figure}

By utilizing the Prompt style multi-task learning method, the abilities of document extraction question answering, document layout understanding, document table understanding, and document classification are integrated into the same model, promoting each other's effectiveness. So a single model can be used to support multiple document tasks simultaneously.


The content of the input model is divided into three parts, namely text, prompt, and image. The text and prompt parts are both composed of text, while the image part cuts the complete image into multiple small pieces, and then extracts the image features of each small piece, transforming them into tokens and inputting them into the model.

Subsequently, after further feature extraction of the pretrain model, the top three linear layers are used for decoding and fusion to obtain the final prediction.

This model structure and multi-task learning scheme can support the vast majority of tasks in the three modalities of text, layout, and image in documents, making it a highly versatile and efficient model scheme.

\subsection{Multiple-stage continue pretrain}

\begin{figure}[h]
\begin{center}
\includegraphics[scale=0.7]{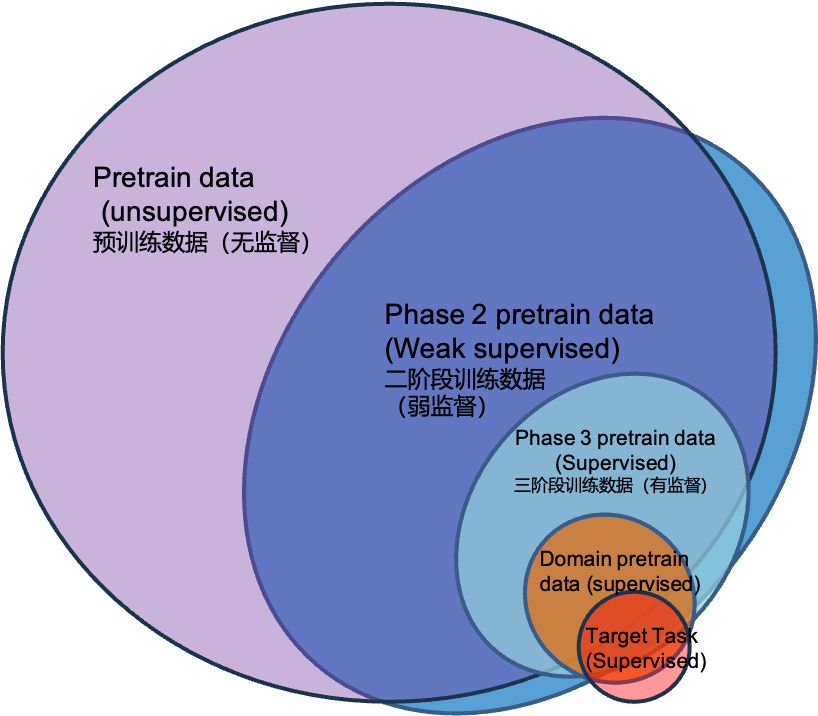}
\end{center}
\caption{Docprompt multi-stage training data}
\end{figure}


In order to better compensate for the gap between pretraining and downstream tasks, and improve the model's zero-shot and few-shot capabilities in document question answering tasks, we need to collect and construct as much rich and diverse document question answering data as possible for further pretraining.

We found that using data which is gradually approaching downstream tasks in multi-stage pretraining can best adapt the model to the data distribution of downstream tasks and improve its performance on downstream tasks.




We have innovatively proposed a multi-stage training scheme for document extraction question answering tasks. Based on the Ernie-layout\cite{peng2022ernielayout} pretrain model, the second stage of pretraining is carried out using billion level weakly supervised document question answering data (as introduced in section 3.1). At this point, the generated model already has most of the basic knowledge and generalization ability, and can be used for zero-shot and few-shot predictions.

Then, in order to better compensate for the gap between weakly supervised training data and real downstream task data. We used a million level open-source document question answering dataset (a publicly available dataset of document question answering tasks collected online) for the third stage of pretraining, further improving the model's generalization ability and answer accuracy. At this point, the output model can better perform predictio in zero-shot and few-shot scenarios.

Finally, based on the specific downstream task scenarios and vertical categories of the customer, we utilize the corresponding vertical category data to further conduct the fourth stage pretraining, and provided it as the basic model to the customer. Customers can directly perform zero-shot inference, or further finetune can be achieved by utilizing customer business data.

\subsection{Understanding-Generation model ensemble}


Currently, there are two paradigms in the field of natural language processing, namely natural language understanding (NLU) and natural language generation (NLG).

The answer to the document question in this article is based on OCR recognition. If there are errors in the text recognized by OCR, these errors are irreparable in the understanding model, and the generated model can precisely solve this problem and achieve OCR error correction. Therefore, we propose a method for integrating understanding models and generating models. This method is compatible with the advantages of understanding models and generating models. By integrating the two paradigms of models, the effectiveness of document question answering tasks can be further improved.

\begin{figure}[h]
\begin{center}
\includegraphics[scale=0.6]{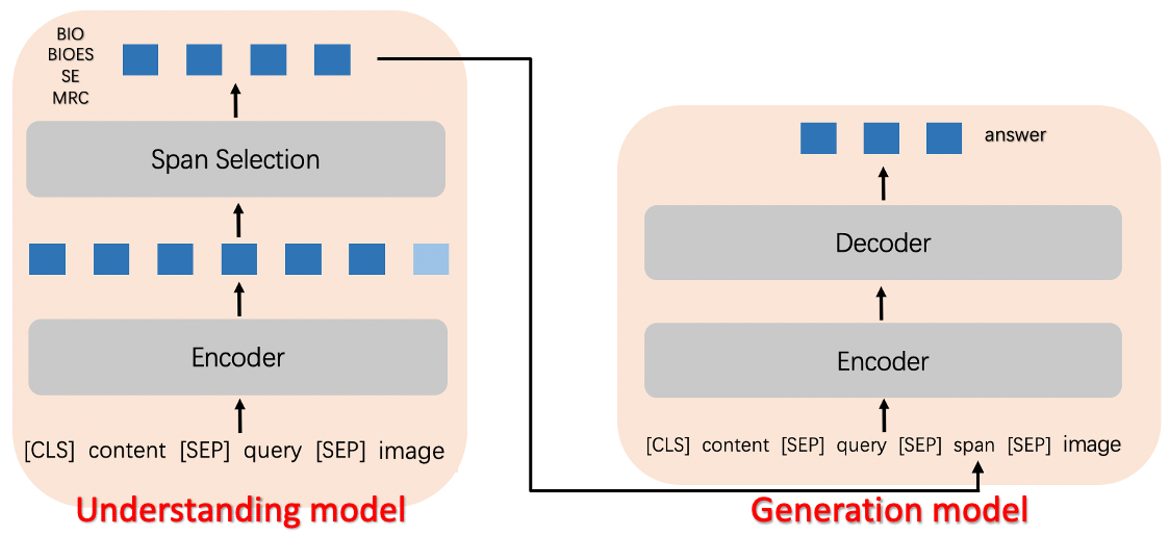}
\end{center}
\caption{Docprompt multi-stage training pipeline}
\end{figure}


We propose NLU2NLG, a solution that combines understanding models and generating models, simulating human left and right brains assisting each other in thinking problems, and achieving knowledge fusion.

\subsubsection{Training method}


Firstly, Train the understanding model separately until loss convergence.
Then, the results of understanding the model are used as a span and concatenated with contextual inputs such as content, prompt, and image as input data for generating the model. The generated model is trained until loss convergence.

\subsubsection{Inference method}


Firstly, we use the understanding model to obtain the predicted results separately. Then, the predicted results of the understanding model are concatenated into the input of the generated model to get the predicted results of the generated model. 
Finally, the predicted results of the generated model will be used as the output results.

\subsubsection{Data construction method}



The text input form for understanding the model side is: [CLS] context [SEP] question [SEP].

The text input form for generating the model side is: [CLS] question [SEP] span [SEP] context [SEP].

Due to the inherent errors in the prediction results of the understanding model, we need to add some perturbations to the training data of the generated model to make it robust to the errors in the understanding model answers. Therefore, the construction scheme for the span in the training data of the generated model is as follows:

For the training set, keep 80\% of the true answers unchanged (ground truth), and randomly perturb the remaining 20\% of the true answers to obtain the span. We used three perturbation methods in total.

\begin{itemize}
	\item Randomly shrink or expand the range of 80\% of the true answers inward or outward;
	\item Replace 10\% of the true answers with the interval of any segment on the current page, where segment refers to the segment granularity interval obtained from OCR parsing of the document.
	\item Replace 10\% of the true answers with any entity interval on the current page, which is derived from the corresponding true answer interval for other questions on the page.
\end{itemize}

\section{Experiments}
\subsection{Model Configurations}


Because Docprompt is based on the Ernie-layout\cite{peng2022ernielayout} model for continuous pretraining, our model configuration follows the Ernie-layout model. Like Ernie-layout, Docprompt also has two sizes of models: base and large. Among them, the Docprompt base model adopts a 12-layer Transformer encoder with 12-head self attention, and the hidden size is 768. Docprompt large model options a 24 layer Transformer encoder with 16 head self attention, hidden size is 1024. Our max sequence length is 512. For documents that exceed the max sequence length, we adopt a sliding window strategy with stride equals 128. The image input will be divided into 7 * 7 blocks and concatenated at the back of the text. Like layoutlmv3, we also add a [CLS] and a [SEP] token at the beginning and end of each text sequence and the bounding box coordinates of it are [0,0,0,0].

\subsection{Training Configurations}

At second stage of pretraining, we used Adam optimizer to train Docprompt with a batch size of 40 for 5 epochs. We use a learning rate of 3e-5 and linearly warm up equals to 10\% steps.

At third stage and fourth stage of pretraining, we used Adam optimizer to train Docprompt with a batch size of 40 for 10 epochs. We use a learning rate of 3e-6 and linearly warm up equals to 10\% steps.

We trained our model with 24 A100 GPUs with 80G memory.

\subsection{Experiment results}

\subsubsection{understanding model experiment results}


We evaluated the effectiveness of our model on four public datasets, including English docvqa\cite{mathew2021docvqa}, Chinese docvqa (insurance question answering competition, where we redivided the test set)\footnote{http://ailab.aiwin.org.cn/competitions/49}, dureadervis\cite{qi-etal-2022-dureadervis}, and websrc\cite{chen2021websrc}.


The results in Table 1 show that according to our multi-stage pretraining scheme, with the continuous pretraining, the effect of our model on the document question answering task is continuously improving, which is much higher than baseline. Our Docprompt single model achieved the first place in the WebSRC leaderboard\footnote{https://x-lance.github.io/WebSRC/index.html}.


This is because in the multi-stage training scheme we designed, the training data in each stage is closer to the data distribution of downstream tasks, gradually making up for the difference between the pre training data distribution and the data distribution of downstream tasks.

\begin{table}[t]
\caption{Docprompt results on document question answering tasks}
\label{results}
\begin{center}
\begin{tabular}{lllll}
\multicolumn{1}{c}{\bf model name} &\multicolumn{1}{c}{\bf Eng DocVQA}  &\multicolumn{1}{c}{\bf ZH DocVQA} &\multicolumn{1}{c}{\bf DuReadervis} &\multicolumn{1}{c}{\bf WebSRC}
\\ \hline \\
matrix &ANLS&ANLS&RougeL&F1\\
\\ \hline \\
LayoutXLM\cite{xu2021layoutxlm}      &-&-&50.94&-\\
MarkupLM\cite{li2022markuplm}      &-&-&-&80.54\\
LayoutLMv2\cite{xu2022layoutlmv2}    &83.48& -&-&-\\
LayoutLMv3\cite{huang2022layoutlmv3}    &83.37& -&-&-\\
\\ \hline \\
Pretrain model\cite{peng2022ernielayout}         &79.55& 70.24&48.52&80.15\\
Second stage pretrain model (Ours)    &81.83(+2.28)&72.06(+1.82)&58.81(+10.29)&83.68(+3.63) \\
Third stage pretrain model  (Ours)    &83.87(+4.32)&72.41(+2.17)&59.21(+10.39)&85.04(+4.89) \\
\\ \hline \\
\end{tabular}
\end{center}
\end{table}


\uline{Note that the score of the pretrain model in table 1 is the result obtained by using the ernie-layout parameter during hot start in the Docprompt model structure (MRC model, BIO, BIOES and SE decoding). This is different from the original usage of Ernie-layout, so the results cannot be aligned with the Ernie-layout paper.} What's more, the results of WebSRC which is reported in this subsection is the results of testset.

\subsubsection{understanding model and generation model ensemble experiment results}


In this section, we report on the effectiveness of integrating understanding models and generating models. The understanding models 1, 2, and 3 shown in the table 2 refer to different versions of Docprompt models trained based on different versions of ERNIE-Layout. We need to ensemble the results of these models.


We concatenate the results from multiple understanding models as multiple spans in the input of the generated model to further improve the final ensemble scores.

The generation model we used here is a new version of ERNIE-Layout, which is encoder-decoder architecture.

\begin{table}[t]
\caption{Understanding model and generating model ensemble results on document question answering tasks}
\label{results}
\begin{center}
\begin{tabular}{lllll}
\multicolumn{1}{c}{\bf model name} &\multicolumn{1}{c}{\bf Eng DocVQA}  
& \multicolumn{3}{c}{\bf WebSRC(devset)}  \\  
\\ \hline \\
matrix                  &ANLS&EM&F1&avg\\
\\ \hline \\
understanding model 1      &85.8&85.21&88.58&86.90\\
understanding model 2      &85.5&83.11&88.64&85.88\\
understanding model 3      &84.53&83.99&89.72&86.85\\
generation model(ERNIE-Layout)      &85.26&-&-&-\\
\\ \hline \\
ensemble model(1 span)        &87.45(+1.65)&85.88(+0.67)&90.47(+1.89)&88.18(+1.28)\\
ensemble model(2 span)        &87.81(+0.36)&85.49&91.59(+1.12)&88.54(+0.36)\\
ensemble model(3 span)        &87.97(+0.16)&86.47(+0.35)&91.58&89.03(+0.49)\\
\\ \hline \\
\end{tabular}
\end{center}
\end{table}



Through case analysis, we found that this method can solve some questions that cannot be solved by the independent generation model. Of course, the introduction of understanding model prediction results also brings a small amount of error information, resulting in a small number of errors.

Finally, we integrate the results of Docprompt understanding model, Ernie-layout understanding model, Ernie-layout generation model and the prediction results of the understanding and generation ensemble model as the final results. 


Our model won the first place in the English DocVQA leaderboard\footnote{https://rrc.cvc.uab.es/?ch=17\&com=evaluation\&task=1}, and broke through the 90 points for the first time.

\subsection{Multi-style decode experiment results}

As in section 3.2.1, we used multi-style decode experiment results to get 3 predictions. After that, we use rule based method to ensemble these predictions to get a better final result. The experiment of this method is shown in table 3. We can see that the results after the ensemble are higher than any of those single decoding method.

\begin{table}[t]
\caption{Multi-style decode experiment results on WebSRC dataset (devset)}
\label{results}
\begin{center}
\begin{tabular}{lllll}
\multicolumn{1}{c}{\bf score} &\multicolumn{1}{c}{\bf EM}  &\multicolumn{1}{c}{\bf ANLS}&\multicolumn{1}{c}{\bf avg}
\\ \hline \\
SE decode      &84.67&87.94&86.31\\
BIO decode      &85.06&88.45&86.76\\
BIOES decode    &85.05&88.43&86.74\\
\\ \hline \\
ensemble result    &85.21&88.58&86.90\\
\\ \hline \\
\end{tabular}
\end{center}
\end{table}

\subsection{Conclusion}



We propose the Docprompt model, which is a document question answering model with powerful zero-shot and few-shot capabilities. We propose an efficient and general weak supervised data construction method for document question answering task, which can greatly save the cost of data annotation and improve the efficiency of data annotation. We also propose a multi-stage, multi task, multi decoding style training scheme, which can make the training data distribution of the model gradually approach the downstream tasks, improve the zero-shot and few-shot effects of the model, and further reduce the training data requirement of downstream tasks. We also propose a new ensemble scheme of understanding model and generation model, which is more accurate and flexible than the vote rule ensemble scheme.

In the future, we want to explore the training method of larger pretraining model in document question answering task and richer multimodal information fusion scheme just like GPT4. We want to integrate the information from the pretraining model of text information, the pretraining model of layout information, and the pretraining model of image information.


\subsubsection*{Acknowledgments}
We are very grateful to Han Liu from IEDAresearch. The Openkv he developed during his work at Baidu was the predecessor of Docprompt, which provided us with valuable inspiration. We also want to thank liya Zhan and yongzhi Liang, they have participated in the Docprompt service framework building and ToB customer project delivery.

\bibliography{reference}
\bibliographystyle{iclr2023_conference}


\end{document}